\documentclass{spieman}  
\usepackage{amsmath,amsfonts,amssymb}
\usepackage{graphicx}
\usepackage{setspace}
\usepackage{tocloft}

\usepackage{url}

\title{Plan in 2D, execute in 3D: An augmented reality solution for cup placement in total hip arthroplasty}

\author[a,*]{Javad Fotouhi}
\author[b]{Clayton P. Alexander, M.D.}
\author[a]{Mathias Unberath}
\author[a]{Giacomo Taylor}
\author[a]{Sing Chun Lee}
\author[a]{Bernhard Fuerst~\footnote{Bernhard Fuerst is now with Verb Surgical Inc.}}
\author[b]{Alex Johnson, M.D.}
\author[b]{Greg Osgood, M.D.}
\author[c]{Russell H. Taylor}
\author[b]{Harpal Khanuja, M.D.}
\author[c,d, $\dagger$]{Mehran Armand}
\author[a,e, $\dagger$]{Nassir Navab}

\affil[a]{Johns Hopkins University, Computer Aided Medical Procedures, 3400 N Charles Street, Baltimore, USA, 21218}
\affil[b]{Johns Hopkins Hospital, Department of Orthopaedic Surgery, 1800 Orleans Street, Baltimore, USA, 21287}
\affil[c]{Johns Hopkins University, Laboratory for Computational Sensing and Robotics, 3400 N Charles Street, Baltimore, USA, 21218}
\affil[d]{Johns Hopkins University Applied Physics Laboratory, 11100 Johns Hopkins Road, Laurel, Maryland, USA, 20723}
\affil[e]{Technische Universit\"at M\"unchen, Computer Aided Medical Procedures, 3 Boltzmannstr, Munich, Germany}

\cftpagenumbersoff{figure}
\cftpagenumbersoff{table} 

\newcommand{\todo}[1]{\textcolor{black}{#1}}
\newcommand{\mat}[1]{\mathbf{#1}}

\begin{document} 
\maketitle

\begin{abstract}   
Reproducibly achieving proper implant alignment is a critical step in total hip arthroplasty (THA) procedures that has been shown to substantially affect patient outcome. 
In current practice, correct alignment of the acetabular cup is verified in C-arm X-ray images that are acquired in an anterior-posterior (AP) view. Favorable surgical outcome is, therefore, heavily dependent on the surgeon's experience in understanding the 3D orientation of a hemispheric implant from 2D AP projection images. 
This work proposes an easy to use intra-operative component planning system based on two C-arm X-ray images that is combined with 3D augmented reality (AR) visualization that simplifies impactor and cup placement according to the planning by providing a real-time RGBD data overlay.
We evaluate the feasibility of our system in a user study comprising four orthopedic surgeons at the Johns Hopkins Hospital, and also report errors in translation, anteversion, and abduction as low as $1.98$\,mm, $1.10^{\circ}$, and $0.53^{\circ}$, respectively. The promising performance of this AR solution shows that deploying this system could eliminate the need for excessive radiation, simplify the intervention, and enable reproducibly accurate placement of acetabular implants.
\end{abstract}


{\noindent \footnotesize\textbf{*}  \linkable{javad.fotouhi@jhu.edu}} 

{\noindent \footnotesize\textbf{$\dagger$} Mehran Armand and Nassir Navab are joint senior authors listed in alphabetical order} 

\begin{spacing}{1.3}   

\section{Introduction}
\label{sect:intro}  

\subsection{Clinical Background} 
In total hip arthroplasty (THA), also referred to as total hip replacement, the damaged bone and cartilage are replaced with prosthetic components. The procedure relives pain and disability with a high success rate. In 2010, there were approximately 330,000 THAs performed in the US. This is projected to increase to 570,000 THAs by 2030~\cite{kurtz2007projections} as younger patients and patients in developing countries are considered for THA. Together with a prolonged life expectancy, the consideration of younger, more active patients for THA suggests that implant longevity is of increasing importance as it is associated with the time to revision arthroplasty~\cite{kurtz2007projections}. The time to repeat surgery is affected by the wear of the implants that is correlated with their physical properties as well as acetabular component positioning. 
Poor placement leads to increased impingement and dislocation that promotes accelerated wear. Conversely, proper implant placement that restores the hip anatomy and biomechanics decreases the risk for dislocation, impingement, loosening, and limb length discrepancy, and thus implant wear and revision rate~\cite{barrack2001virtual, charnley1973nine, d2000effect, scifert1998finite, yamaguchi2000spatial}. Steps to ensure accuracy and repeatability of acetabular component positioning are therefore essential. Due to the large volume of THA procedures, small but favorable changes to the risk-benefit profile of this procedure enabled by improved implant positioning will have a significant impact on a large scale.

Unfortunately, optimal placement of the acetabular component is challenging due to two main reasons. First, the ideal position of the implant with respect to the anatomy is unknown; yet, a general guideline exists~\cite{lewinnek1978dislocations} and is widely accepted in clinical practice. This guideline suggests abduction and anteversion angles of the hip joint measured with respect to bony landmarks defining the so-called safe zone, that is indicative for an acceptable outcome. 
Recent studies suggest that an even narrower safe zone may be necessary to minimize the risk of hip dislocation~\cite{elkins20152014,danoff2016redefining}. Defining the ideal implant position is not as straight-forward as the definition of a range of abduction and anteversion angles when considering a large population~\cite{esposito2015cup}. A static definition of the safe zone seems even more prone to error when considering that the position of the pelvis varies dramatically from supine to sitting to standing posture among individuals~\cite{ digioia2006functional, zhu2010quantification}.

Second, even if a clinically acceptable safe zone is known it is questionable whether surgeons are, in fact, able to accomplish acetabular component placement within the suggested margin~\cite{danoff2016redefining}. In light of previous studies that report mal-positioning of up to 30$\%$ to 75$\%$~\cite{ callanan2011john,  bosker2007poor, saxler2004accuracy} when free-hand techniques are used, addressing this challenge seems to be imperative. 

Most computer-assisted methods consider the direct anterior approach (DAA) to the hip for THA, as it allows for convenient integration of intra-operative fluoroscopy to guide the placement of the acetabular component\todo{~\cite{anterior2009outcomes}}. The guidance methods reviewed below proved effective in reducing outliers and variability in component placement, which equates to more accurate implant positioning~\cite{domb2014comparison, dorr2007precision, moskal2011acetabular, murphy2006tha}.

\subsection{Related work:}  
External navigation systems commonly use certain points on the anatomy of interest, as decided by the surgeon, and conform to a "map" of the known morphology of the anatomy of  interest. \todo{Despite the fact that THA commonly uses X-ray images for navigation and pre-operative patient CT may not be available, several computer assisted THA solutions suggest planning the desired pose of the acetabular component pre-operatively on a CT scan of the patient~\cite{ digioia2000surgical, jaramaz20062d}. Pre-operative CT imaging allows planning of the implants in 3D, automatically estimating the orientation of the natural acetabular opening, and predicting the appropriate size of the cup implant~\cite{nikou1999pop}.}


\todo{Navigation-based THA with external trackers are performed based on pre-operative patient CT, or image-less computer assisted approaches. The planning outcome in a CT-based navigation approach is used intra-operatively with external optical navigation systems to estimate the relative pose of the implant with respect to the patient anatomy during the procedure. Tracking of the patient is commonly performed using fiducials that are drilled into the patient's bones. Registration of the pre-operative CT data to the patient on the surgical bed is performed by manually touching anatomical landmarks on the surface of the patient using a tracked tool~\cite{digioia2000surgical}. In addition to the paired-point transformation estimated by matching the few anatomical landmarks, several points are sampled on the surface of the pelvis and matched to the segmentation of the pelvis in the CT data~\cite{leenders2002reduction}. CT-based navigation showed statistically significant improvement in orienting the acetabular component and eliminating malpositioning, while resulting in increased blood loss, cost, and time for surgery~\cite{widmer2004joint, haaker2007comparison}. Combined simultaneous navigation of the acetabulum and the femur was used in 10 clinical tests where the surgical outcome based on post-operative imaging showed $2.98$\,mm and $4.25^{\circ}$ error in cup position and orientation, respectively~\cite{sato2000intraoperative}.}

\todo{Image-less navigation systems do not require any pre-operatively acquired radiology data. In this method, the pelvic plane is located in 3D by only identifying anatomical landmarks on the surface of the patient using a tracked pointer reference tool and optically visible markers attached to the patient~\cite{sarin2004non}. This approach showed improvement in terms of cup positioning~\cite{kalteis2006imageless}. However, few number of samples points for registration as well as pelvis tilts resulted in unreliable registration~\cite{lin2011limitations}.}


Robotic systems are developed to provide additional confidence to the surgical team in placing implants during THA~\cite{taylor1994image, sugano2013computer}. In a robotic system, pins are implanted into the patient's femur prior to acquiring a pre-operative CT scan. After the surgeon has performed the planning on the CT data, the robot is introduced into the operating room. To close the registration loop between patient, robot, and CT volume, each pre-operatively implanted pin is touched by the robot with manual support. \todo{To eliminate the need for fiducial implantation, registration is either achieved by selecting several points on the surface of the bone using a digitizer and using an iterative closest point algorithm to perform registration to patient CT data~\cite{nakamura2009robot}, or by using intra-operative C-arm fluoroscopy and performing 2D/3D registration between the X-ray image and CT volume~\cite{yao2000ac}.} \todo{After registering the pre-operative CT data to patient}, the robot assists the surgeon in placing the femoral stem \todo{and the acetabular component} according to the planning. The outcome of 97 robot-assisted THA procedures indicates performance similar to the conventional technique~\cite{schulz2007results}; However, in some cases additional complications such as nerve damage, post-operative knee effusion, incorrect orientation of the acetabular component, and deep reaming resulting in leg length discrepancy were reported when the robotic system was used. \todo{To assist the surgeon in placing implants for joint replacement procedures, haptic technology was integrated into robotic solutions to maintain the orientation of the cup according to pre-operative planning and control the operator's movement~\cite{nawabi2013haptically}.}

If pre-operative CT is available, intensity-based 2D/3D registration can be used to evaluate and verify the positioning of the acetabular component post-operatively~\cite{jaramaz20062d}. This is done by recovering the spatial relation between a post-operative radiograph (2D) and the pre-operative patient CT data (3D), followed by a registration of the 3D CAD model of the component to the 2D representation of the cup in the post-operative radiograph. To overcome the large variability in individual patient pelvic orientations, and to eliminate the need for pre-operative 3D imaging, the use of deformable 2D/3D registration with statistical shape models was suggested~\cite{zheng2009statistically}.


Aforementioned solutions perform well but require pre-operative CT \todo{which increases the time and cost for surgery and requires intra-operative registration to the patient~\cite{widmer2004joint, sugano2013computer}.} Zheng et al. proposed a CT-free approach for navigation in THA~\cite{zheng2002hybrid}. \todo{The method relies on tracking of the C-arm, surgical instruments used for placing femoral and acetabular components, as well as the patient's femur and pelvis using an external navigation system. Multiple stereo C-arm fluoroscopy images are acquired intra-operatively. Anatomical landmarks are then identified both in these X-ray images as well as percutaneously using a point-based digitizer. Thanks to the tracking of the C-arm, the relative pose between the X-ray images are known, therefore the anatomical landmarks are triangulated from the images and reconstructed in 3D. These points are used later to define the anterior pelvic plane and the center of rotation for the acetabulum. After estimating the pelvis coordinate frame, the impactor is moved by the surgeon until the cup is at the correct alignment with respect to a desired orientation in the anterior pelvic plane coordinate frame.} This work reported sub-degree and sub-millimeter accuracy in antetorsion, varus/valgus, and leg length discrepancy. Later, this system was tested in 236 hip replacement procedures, where a maximum of $5^{\circ}$ inclination error, and $6^{\circ}$ anteversion error were observed~\cite{grutzner2004c}.


\todo{The state of the art approaches that provide guidance using image-less or image-based methods have certain drawbacks. Image-less methods require complex navigation and may provide unreliable registration~\cite{lin2011limitations}. Image-based solutions rely on pre-operative CT scans or intra-operative fluoroscopy and often use external navigations systems for tracking~\cite{xu2014computer,reininga2010minimally}.} Systems based on external navigation are expensive and increase the operative time due to the added complexity. \todo{Use of pre-operative CT scans} increases the radiation exposure and cost to the patient. Moreover, many of the methods used for registering CT to patient seek to solve ill-posed problems that require manual interaction either for initialization or landmark identification and, thus, disrupt the surgical workflow. Manual annotations can take between 3 to 5 minutes during the intervention for \emph{each} image registration~\cite{digioia2000surgical}. Although proven beneficial for the surgical outcome, neither of these costly and labor-intensive navigation techniques were widely adopted in clinical practice.

Partly due to above drawbacks, surgeons who use the DAA often rely solely on fluoroscopic image guidance\todo{~\cite{anterior2009outcomes,barrett2013prospective}}. These images, however, are a 2D representation of 3D reality and have inherent flaws that complicate the assessment. The challenges include finding the true anterior pelvic plane as well as eyeballing acetabular component position by eye on the image. Therefore, a technique that provides a quantitative and reliable representation of the pelvis and acetabular component intra-operatively without increasing neither radiation dose or cost while largely preserving the procedural workflow is highly desirable.

\subsection{Proposed solution:} 
\begin{figure*}
  \centering
  \includegraphics[width=\textwidth]{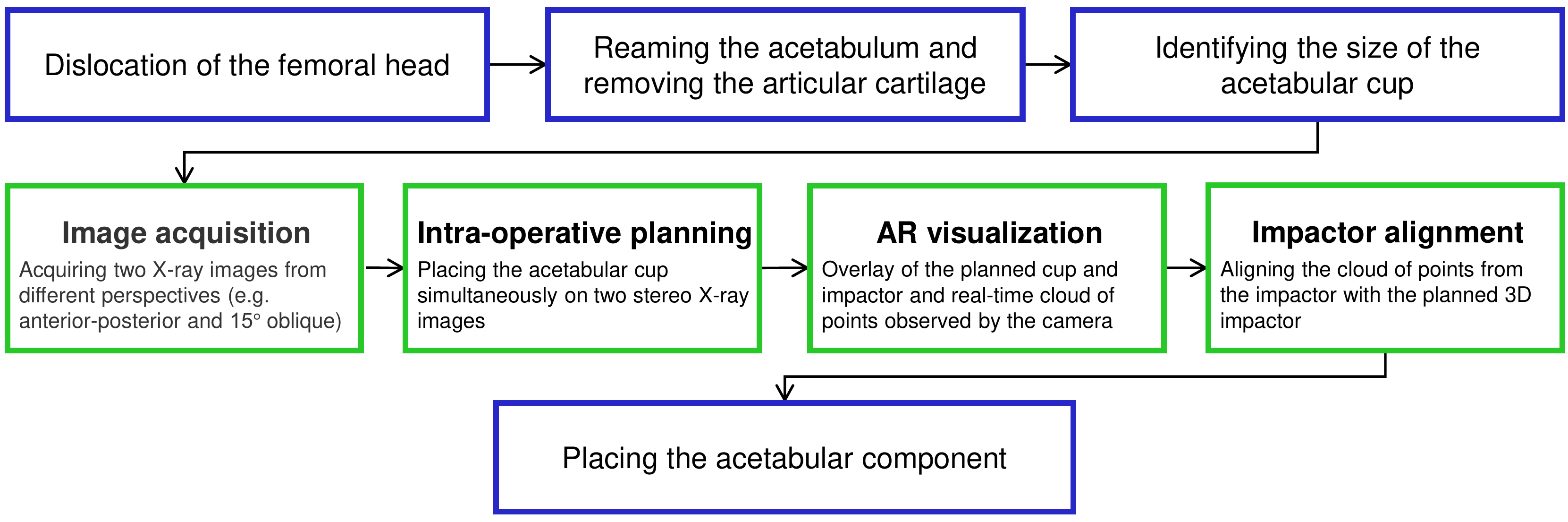}
  \caption{After the femoral head is dislocated, the size of the acetabular implant is identified based on the size of the reamer. Next, two C-arm X-ray images are acquired from two different perspectives. While the C-arm is repositioned to acquire a new image, the relative poses of the C-arm are estimated using the RGBD camera on the C-arm and a visual marker on the surgical bed. The surgeon then plans the cup position intra-operatively based on these two stereo X-ray images simultaneously. Next, the pose of the planned cup and impactor are estimated relative to the RGBD camera. This pose is used to place the cup in a correct geometric relation with respect to the RGBD camera and visualize it in an AR environment. Lastly, the surgeon observes real-time optical information from the impactor, and aligns it with the planned impactor using the AR visualization. The green boxes in this figure highlight the contributions of this work.}
    \label{fig:workflow}
\end{figure*}

This work proposes an augmented reality (AR) solution for intra-operative guidance of THA \todo{using DAA where the C-arm is kept in place until the correct alignment of the acetabular cup is confirmed~\cite{slotkin2015accuracy,masonis2008safe}}. With the proposed solution, the surgeon first plans the position of the acetabular cup intra-operatively based on two fluoroscopy images that are acquired after the dislocation of the femoral head and the reaming of the acetabulum are completed. The orientation of the cup in the X-ray images could be either automatically preset based on desired angles relative to the APP plane (or other known pelvic coordinate frames), or be adjusted by the surgeon.
Once the desired pose of the acetabular cup is estimated relative to the C-arm, we use optical information from the co-calibrated RGBD camera that is mounted on the C-arm to provide an AR overlay~\cite{lee2016calibration,fotouhi2016interventional,fischer2016preclinical} that enables placement of the cup according to the planning. As the cup is not visible in RGBD, we exploit the fact that the acetabular cup is placed using an impactor that is rigidly attached to the cup and is well perceived by the RGBD camera. For accurate cup placement, the surgeon aligns the optical information of the impactor (a cloud of points provided by the RGBD camera) with the planned virtual impactor-cup, that are visualized simultaneously in our AR environment. A schematic of the proposed clinical workflow is shown in Fig.~\ref{fig:workflow}.

\section{Methodology}
\label{sect:method}

The AR environment for THA requires a co-calibrated RGBD-C-arm (Sec.~\ref{subsec:cocalibration}). Whenever the C-arm is re-positioned, the RGBD camera on the C-arm tracks and estimates C-arm relative extrinsic parameters (Sec.~\ref{subsec:markertracking}). During the intervention, two X-ray images are recorded at different poses together with the respective extrinsic parameters, and are used for intra-operative planning of the component (Sec.~\ref{subsec:twoViewplannning}). Lastly, an AR environment is provided for the placement of the cup that comprises of surface meshes of a virtual cup and impactor displayed in the pose obtained by intra-operative planning, overlaid with the real-time cloud of points from the surgical site acquired by the RGBD camera (Sec.~\ref{subsec:AR}).

\subsection{Co-calibration of the RGBD-C-arm imaging devices} 
\label{subsec:cocalibration}
\begin{figure*}
  \centering
  \includegraphics[width=\textwidth]{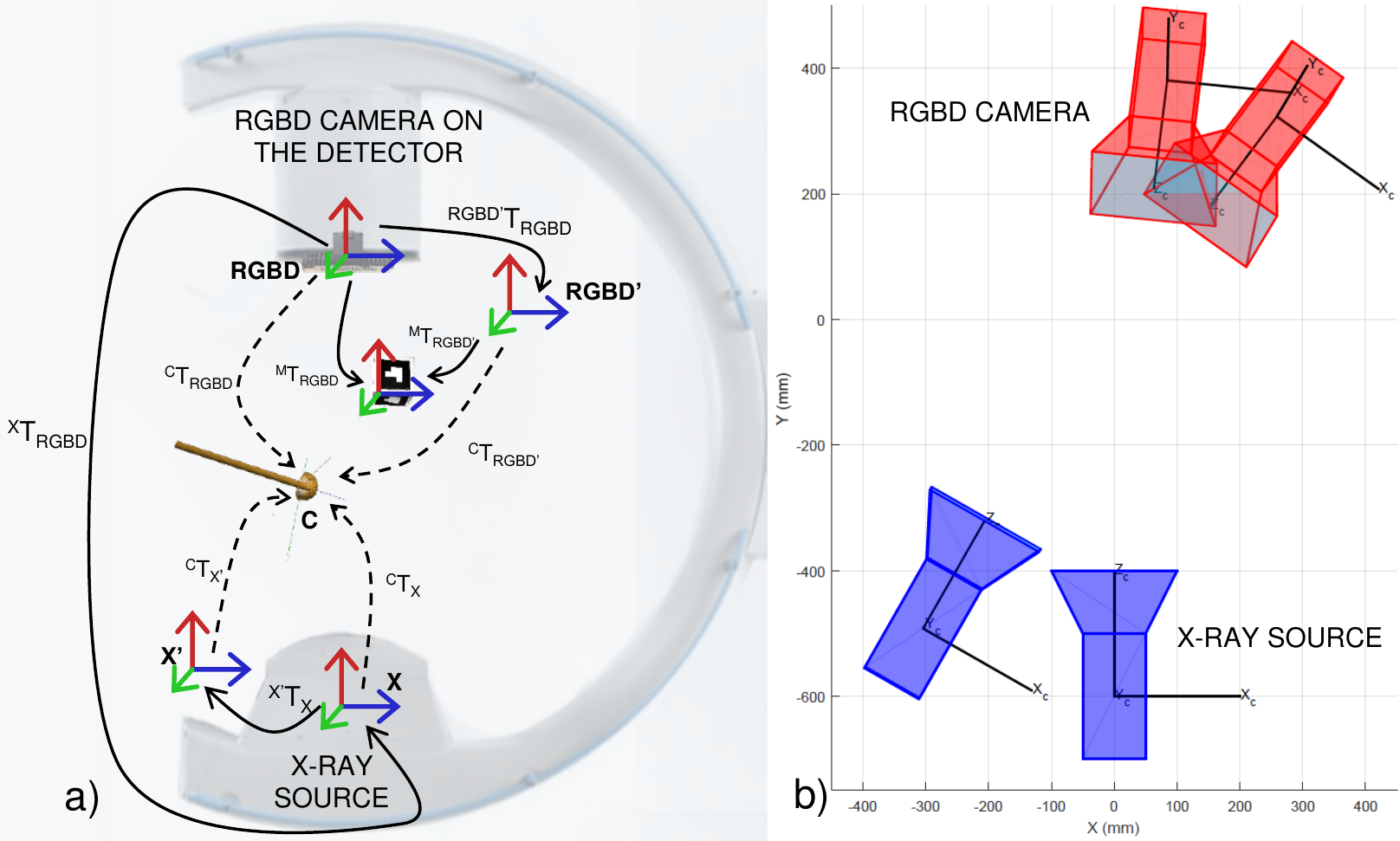}
  \caption{In the transformation chain of the RGBD-C-arm system for THA \textbf{(a)}, the RGBD, X-ray, visual marker, and acetabular cup coordinate frames are denoted as \textbf{RGBD}, \textbf{X}, \textbf{M}, and \textbf{C}, respectively. In an offline calibration step, the extrinsic relation between the RGBD and X-ray ($^{\text{X}}\mat{T}_\text{RGBD}$) is estimated. Once this constant relation is known, the pose of the X-ray source can be estimated for every C-arm re-positioning \textbf{(b)} by identifying displacements in the RGBD camera coordinate frame.}
  \label{fig:transformations}
\end{figure*}

The co-calibration of the RGBD camera and the X-ray imaging devices is performed using a multi-modal checkerboard pattern. In this hybrid checkerboard pattern, each black square is backed with a radiopaque thin metal square of the same size~\cite{fotouhi2017can}. Calibration data is acquired by simultaneously recording RGB and X-ray image pairs of the checkerboard at various poses. Next, we estimate the intrinsic parameters for both the RGB channel of the RGBD sensor as well as for the X-ray imaging device. Using these intrinsic parameters, we estimate the 3D locations of each checkerboard corner, $\mat{O}_{\text{CH}_\text{RGB}}$ and $\mat{O}_{\text{CH}_\text{X}}$ in the RGB and X-ray coordinate frames, respectively. The stereo relation between the X-ray and RGB imaging devices $^\text{X}\mat{T}_\text{RGB}$ is then estimated via least squares minimization:
\begin{equation}
\label{eq:rgbTOX}
\min_{^\text{X}\mat{T}_\text{RGB}} ~~ \| ^\text{X}\mat{T}_\text{RGB} \mat{O}_{\text{CH}_\text{RGB}} - \mat{O}_{\text{CH}_\text{X}} \|_2^2 \,.
\end{equation}
The stereo relation between the RGB and IR (or depth) channel of the RGBD sensor is provided by the manufacturer. To simplify the notation, we use $^\text{X}\mat{T}_\text{RGBD}$ which embeds the relation between the RGB, depth, and the X-ray imaging devices. 
\todo{We assume that both extrinsic parameters between X-ray and RGBD and the intrinsic parameters of the X-ray remain constant, however, both quantities are subject to minor change while the C-arm rotates to different angles, an observation that is further discussed in Sec.~\ref{Sec::discuss}.} Fig.~\ref{fig:transformations}-a illustrates the spatial relation between the RGBD camera and the X-ray source. 
%

\subsection{Vision-based estimation of C-arm extrinsic parameters}
\label{subsec:markertracking}
The stereo relation between C-arm X-ray images acquired at different poses are estimated by first tracking visual markers in the RGBD camera coordinate frame, and then transforming the tracking outcome to the X-ray coordinate frame:
\begin{equation}
^{\text{X}^{\prime}}\mat{T}_\text{X} = ^{\text{RGBD}^{\prime}}\mat{T}_{\text{X}^{\prime}}^{-1} ~ ^\text{M}\mat{T}_{\text{RGBD}^{\prime}}^{-1} ~ ^{\text{M}}\mat{T}_\text{RGBD} ~ ^{\text{RGBD}}\mat{T}_\text{X} \,,
\end{equation}
where $^{\text{RGBD}^{\prime}}\mat{T}_{\text{X}^{\prime}} = ^{\text{RGBD}}\mat{T}_\text{X}$ due to the rigid construction of the RGBD camera on the C-arm gantry. In  Fig.~\ref{fig:transformations}-b the rigid movement of X-ray source with the RGBD camera origin are shown for an arbitrary C-arm orbit.

\subsection{Intra-operative planning of the acetabular cup on two X-ray images}
\label{subsec:twoViewplannning}

Planning of the acetabular component is performed in a user interface where the cup could be rotated and translated by the surgeon in 3D with six degrees of freedom (DOF) rigid parameters, and is forward projected ($p_{c_v}$ and ${p^{\prime}}_{c_v}$) onto the planes of the two X-ray images acquired from different perspectives:

\begin{align}
	p_{c_v} &= \text{K} ~ \text{P} ~ ^\text{C}\mat{T}_{\text{X}}^{-1}  ~ ^{\text{C}_v}\mat{T}_{\text{W}} \,,\nonumber \\
	{p^{\prime}}_{c_v} &= \text{K}^{\prime} ~ \text{P} ~ ^\text{C}\mat{T}_{\text{X}^{\prime}}^{-1}  ~ ^{\text{C}_v}\mat{T}_{\text{W}} \,,
\end{align}
where $\text{K}$ and $\text{K}^{\prime}$ are the intrinsic perspective projection parameters for each C-arm image, $\text{P}$ is a projection operator, and $^{\text{C}_v}\mat{T}_{\text{W}}$ is the position of vertex $v$ of the cup in the world coordinate frame. Relying on two X-ray views not only provides the ability to plan the orientation of the acetabular component such that it is aligned in two images but, more importantly, also allows adjusting the depth of the cup correctly, which is not possible when a single X-ray image is used. It is worth mentioning, that the size of the acetabular cup does not require adjustment but is known at this stage of the procedure as it is selected to match the size of the reamer. 

In addition, if the desired orientation of the cup is known relative to an anatomical coordinate frame (e.g. APP plane), and an X-ray image is acquired from a known perspective in relation to that anatomical frame (e.g. AP view), then the orientation of the cup could be automatically adjusted for the user (equivalent to presetting the orientation in $^\text{C}\mat{T}_{\text{X}}$). It is worth emphasizing that in several image-guided orthopedic procedures, X-ray images are frequently acquired from the AP view.

\begin{figure*}
  \centering
  \includegraphics[width=\textwidth]{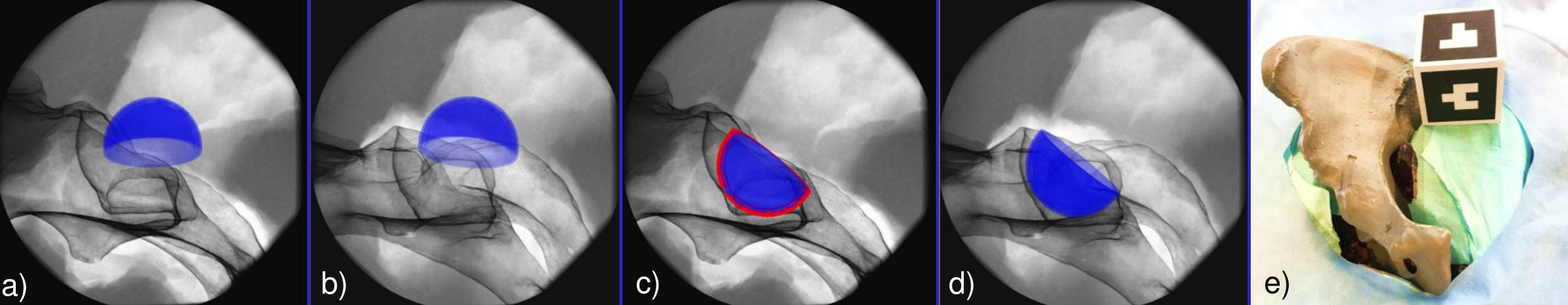}
  \caption{The acetabular component is forward projected from an initial 3D pose onto the respective X-ray image plane ~\textbf{(a-b)}. The surgeon moves the cup until satisfied with the alignment in both views~\textbf{(c-d)}. The X-ray images shown here are acquired from a dry pelvis phantom encased in gelatin. A cubic visual marker is placed near the phantom but outside the X-ray field of view to track the C-arm~\textbf{(e)}.}
  \label{fig:planning}
\end{figure*}

The transparency of the cup is adjusted by the surgeon in the user interface such that the ambiguity between the front and the back of the cup is optimally resolved. Lastly, the contours around the edge of the cup are estimated and visualized by thresholding the dot product of the unit surface normal $\mat{n}_v$ and the intersecting ray $\mat{r}_v$: 

\begin{equation}
\mid \mat{r}_v~.~\mat{n}_v \mid ~ < ~ \tau \,.
\end{equation}
The planning of an acetabular cup based on two X-ray images is shown in Fig.~\ref{fig:planning}.  

\subsection{Augmented-reality visualization} 
\label{subsec:AR}
Once the desired cup position is known, guidance of the cup placement using an impactor with an AR visualization is needed to ensure a positioning in agreement with the planning. To construct the AR environment, we first estimate the pose of the RGBD sensor relative to the planned cup as
\begin{equation}
^\text{C}\mat{T}_\text{RGBD} = ^\text{C}\mat{T}_\text{X} ~ ^\text{X}\mat{T}_\text{RGBD} \,.
\end{equation}
Within the AR environment, we then render a 3D mesh of the cup and impactor superimposed with the real-time cloud of points observed by the camera, all in the \textbf{RGBD} coordinate frame. In the interventional scenario, the acetabular cup is hidden under the skin, and only the impactor is visible. Therefore, the surgeon will only align the cloud of points from the impactor, a cylindrical object, with the 3D virtual representation of the planned impactor.

\begin{figure*}
  \centering
  \includegraphics[width=\textwidth]{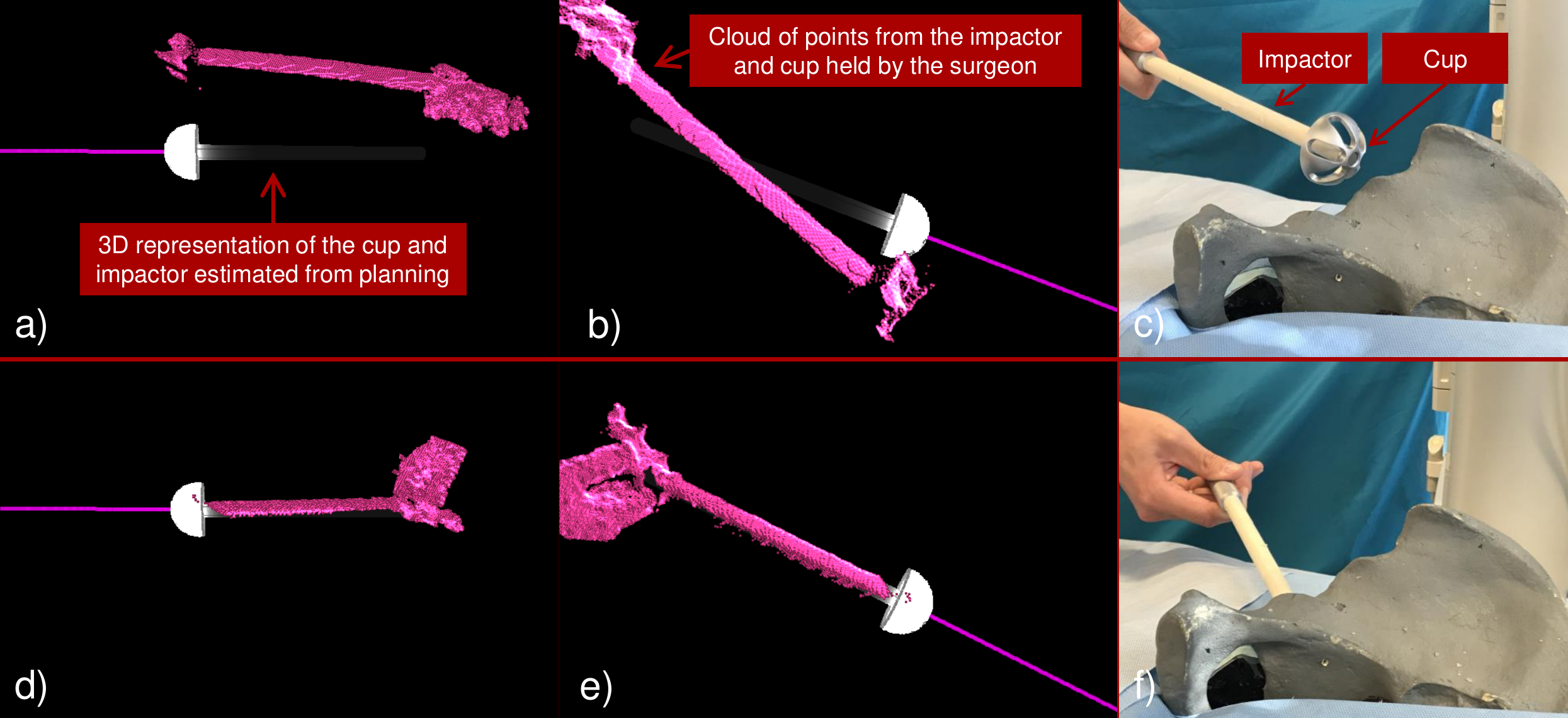}
  \caption{Multiple virtual perspectives of the surgical site are shown to the surgeon~\textbf{(a-b)} before the cup is aligned~\textbf{(c)}. The impactor is then moved by the user until it completely overlaps with the virtual planned impactor~\textbf{(d-f)}.}
  \label{fig:ARvisualization}
\end{figure*}

Ambiguities in the AR environment, among others occlusions or the rendering of a 3D scene in a 2D display, are eliminated by showing different perspectives of the scene simultaneously. Thus, it is ensured that the surgeon's execution fully matches the planning once alignment of the current cloud of points of the impactor and the planned model is achieved in all perspectives. We provide an intuitive illustration of these relations in Fig.~\ref{fig:ARvisualization}.

\todo{In order to solely visualize the moving objects (e.g. the surgeon's hands and tools), background subtraction of point clouds is performed with respect to the first static frame observed by the RGBD camera. It is important to note that in an image-guided DAA procedure, most tools other than the impactor, such as retractors, are removed prior to placing the acetabular component, therefore important details on the fluoroscopy image are not occluded.}

\section{Evaluation and Results}
\subsection{System Setup}

\todo{In DAA for THA the detector is commonly positioned above the surgical bed. This orientation of the C-arm machine is considered to reduce the scattering to the surgical crew. Therefore, we modified the C-arm machine by mounting the RGBD camera near the detector plane of the C-arm which then allows the detector to remain above the bed. The mount for the RGBD camera extends out from the C-arm detector for nearly $5.00$\,cm in XY direction (Z being the principal axis of the X-ray camera) and is screwed to a straight metal plate which is rigidly tied to the image intensifier. Considering that the RGB camera is used for pose estimation of the C-arm scanner while the depth camera is used for point cloud observation, the RGBD camera needs to be angled such that a maximum of the surgical site is visible in both RGB and depth camera views. The RGBD sensor is, therefore, angled such that it has a direct view onto the surgical site such that the principal axis of the camera is close to the iso-center of the C-arm.}

\todo{The impactor used for testing is a straight cylindric acetabular trialing from Smith and Nephew.} For intra-operative X-ray imaging, we use an Arcadis Orbic 3D C-arm (Siemens Healthcare GmbH, Forchheim, Germany) with an iso-centeric design and an image intensifier. The RGBD camera is a short range Intel RealSense SR300 (Intel Corporation, Santa Clara, CA) which combines depth sensing with HD color imaging. Data transfers from C-arm and the RGBD camera to the development PC are done via Ethernet and powered USB 3.0 connections, respectively.

The AR visualization is implemented as a plug-in application in ImFusion Suite using the ImFusion software development kit~\footnote{\url{http://imfusion.de/products/imfusion-suite}}. We use ARToolkit~\footnote{\url{https://artoolkit.org/}} for visual marker tracking~\cite{kato1999marker}. 

\subsection{Experimental Results}

\paragraph{Stereo co-calibration of the RGBD and X-ray cameras:} Offline stereo co-calibration between the X-ray source and the RGBD camera using 22 image pairs yields a mean reprojection error of $1.10$ pixels. Individual mean reprojection errors for X-ray and RGBD cameras are $1.46$ and $0.74$ pixels, respectively.

\paragraph{Accuracy in tracking X-ray poses:} Tracking accuracy is computed by acquiring X-ray images from a phantom with several radiopaque landmarks and measuring the stereo error between the corresponding landmark points in different images. 

\begin{figure*}
  \centering
  \includegraphics[width=0.8\textwidth]{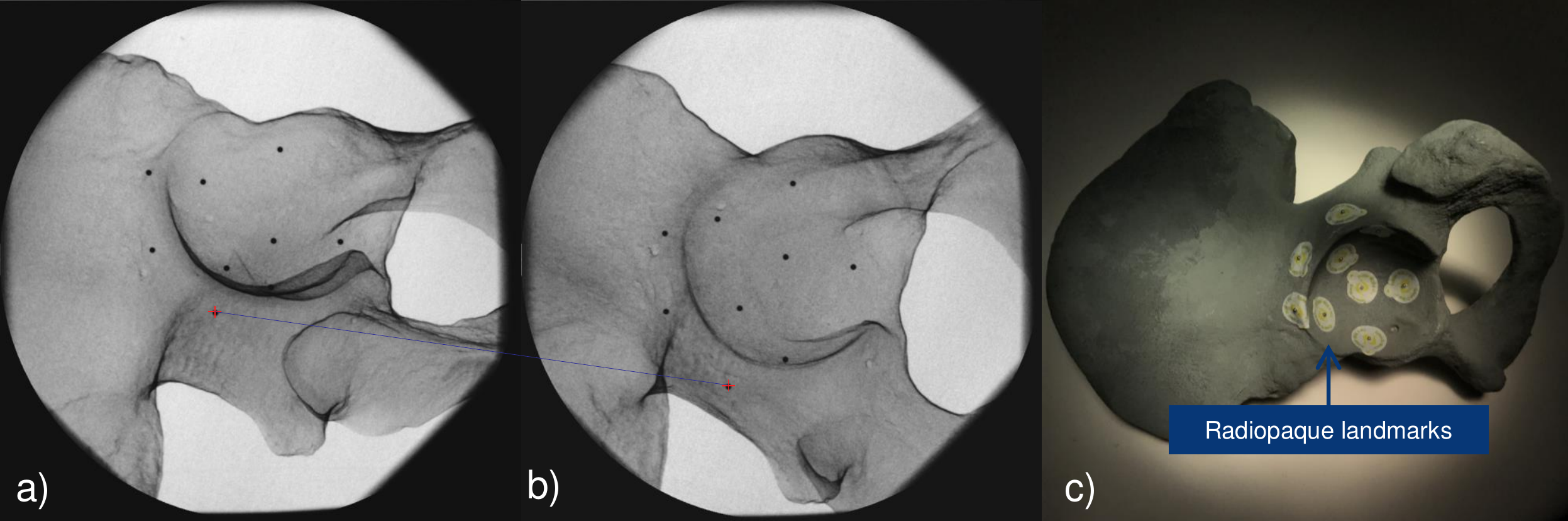}
  \caption{The geometric error is measured using radiopaque bbs viewed in the stereo X-ray images~\textbf{(a-b)}. The blue line highlights a pair of corresponding bbs in the two images. The phantom is shown in~\textbf{(c)}.}
  \label{fig:bbphantom}
\end{figure*}

The phantom is constructed by attaching 9 radiopaque mammography skin markers (bbs) with diameters of $1.5$\,mm, inside and near the acetabulum on a pelvis model as shown in Fig.~\ref{fig:bbphantom}. Next, we acquired 11 X-ray images from $-50^{\circ}$ to $+50^{\circ}$ along C-arm oblique rotation, and 9 X-ray images from $-40^{\circ}$ to $+40^{\circ}$ on the cranial/caudal direction, with intervals of $10^{\circ}$. In the planning software, we placed a virtual sphere with the same diameter as the bbs on each of the bb landmarks, and measured the distance of the bb in the second image to the epipolar line from the center of the corresponding virtual sphere in the first image. The error distance is measured as $7.58 \pm 3.02$ pixels\footnote{values reported as mean $\pm$ standard deviation} \todo{in an X-ray image with pixel size of $1024 \times 1024$ and pixel spacing of $0.22$\,$\frac{\text{mm}}{\text{pixel}}$.} In addition, we acquired a cone beam CT (CBCT) scan of the phantom and measured a root mean square error of $1.37$\,mm between the bbs in the CT and those reconstructed using two X-ray images.   

\begin{figure*}
  \centering
  \includegraphics[width=0.8\textwidth]{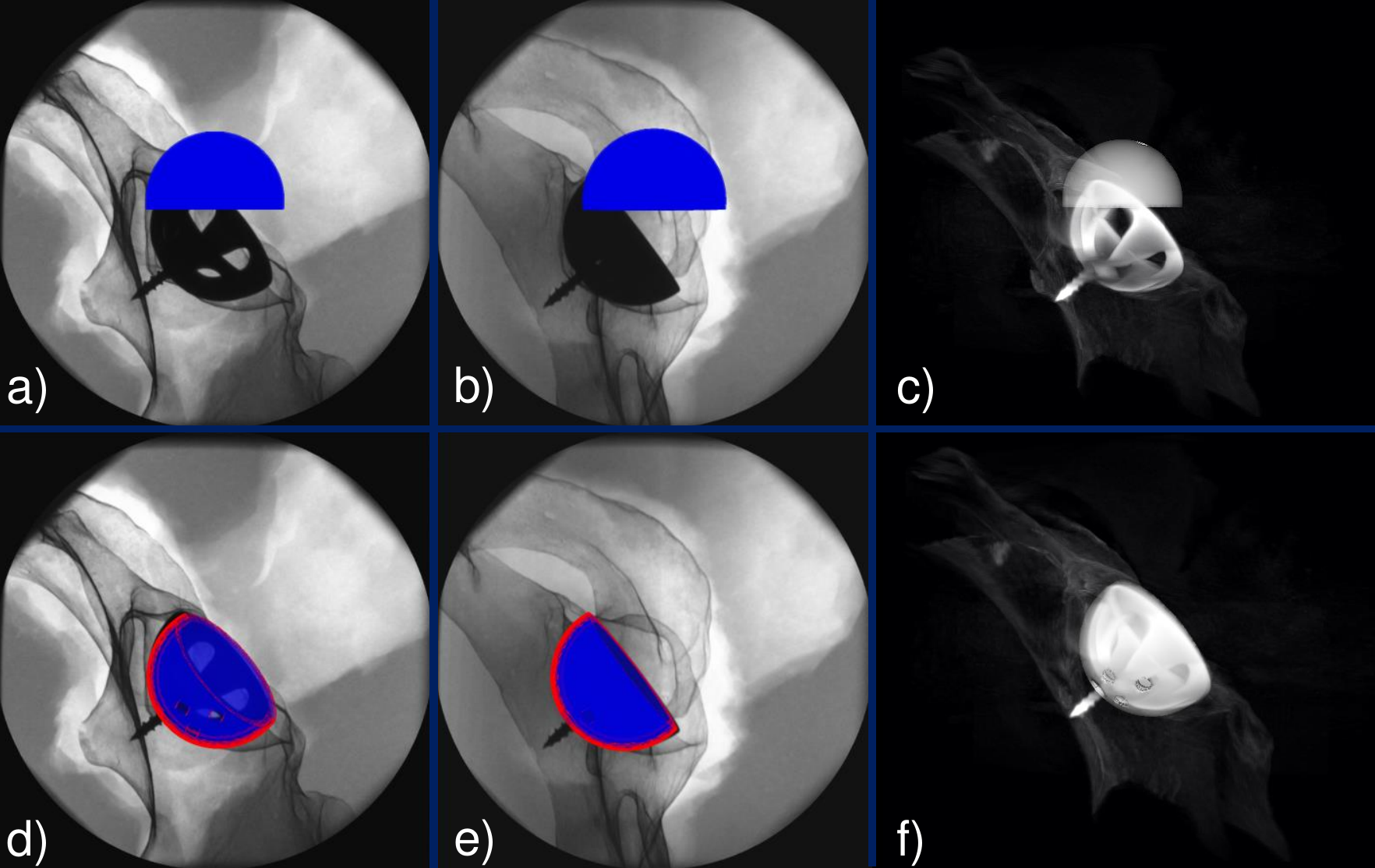}
  \caption{An implant cup is placed inside the acetabulum and two X-ray images and a CBCT scan are acquired using the C-arm. \textbf{(a-b)} and \textbf{(c)} are the X-ray and CBCT images before planning. \textbf{(d-f)} show the overlay of the real and virtual cup after proper alignment.}
    \label{fig:cupOncup}
\end{figure*}

\paragraph{Planning accuracy in placing the acetabular component using two views:} 
In order to measure 3D errors and ensure precise placement of the cup in two X-ray images during planning, we construct a dry phantom where an implant cup is screwed into the acetabulum. Therefore, the desired implant cup placement is well visible in the X-ray images and serves as a reference.
We perform experiments where a virtual cup with the same size of the implant, shown in Fig.~\ref{fig:cupOncup}, must be aligned precisely with the cup implanted a priori that is visible in the X-ray images. To evaluate the 3D error, we acquire a CBCT scan of the phantom and measure the error between the planning outcome and the ground-truth pose. This yields a mean translation error of $1.71$\,mm, and anteversion and abduction errors of $0.21^{\circ}$ and $0.88^{\circ}$, respectively.   

\paragraph{Pre-clinical feasibility study of acetabular component planning using stereo X-ray imaging:} In image-guided DAA hip arthroplasty, the proper alignment of the acetabular component is frequently inferred from AP X-ray images\todo{~\cite{ji2016fluoroscopy}}. Thus, the accuracy in estimating the 3D pose based on a single 2D image heavily depends on the surgeon's experience. In this experiment, we seek to demonstrate the clinical feasibility of our solution that is based on stereo X-ray imaging, and compare the outcome to \todo{image-guided DAA solutions that only use AP X-ray images for guidance. We refer to the latter as} "classic DAA". \todo{Although the use of a single AP radiograph and the anterior pelvic plane coordinate system have certain drawbacks, it is the frame of reference that is most commonly used in computer-assisted THA solutions~\cite{rousseau2009optimization}. While  there may be alternatives (e.g. coronal plane), the use of anterior pelvic plane as the frame of reference will enable direct comparison with the current literature.}

We conduct a pre-clinical user study where medical experts use the planning software to place acetabular cups on simulated stereo X-ray images. These results are then compared to conventional AP-based method considering orientational error in abduction and anteversion.

For the  purpose of the user study, simulated X-ray images or so-called digitally reconstructed radiographs (DRR) are produced from a cadaver CT data. We generate 21 DRRs from the hip area, starting at $-45^{\circ}$ and ending at $+45^{\circ}$ with increments of $+4.5^{\circ}$ on the orbital oblique axis of the C-arm, where $0^{\circ}$ refers to an AP image. Each time the users are given a randomly selected DRR together with the DRR corresponding to the AP plane, and are expected to place the acetabular cup such that it is properly aligned in both views. 

As the spatial configuration of the DRRs are known relative to the APP plane, we are able to compute the correct rotation of the acetabular component, and preset this orientation for the cup in the planning software. \todo{This can occur when an AP image is acquired during the intervention and the desired orientation of the component is known relative to the anterior pelvic plane which allows locking the DOF for rotational parameters.} When the orientation is preset, the user only has to adjust a translational component, substantially reducing the task load. Presetting the orientation of the cup is evidently only possible if the X-ray pose is known relative to the APP or the AP image. 

\begin{figure*}
  \centering
  \includegraphics[width=\textwidth]{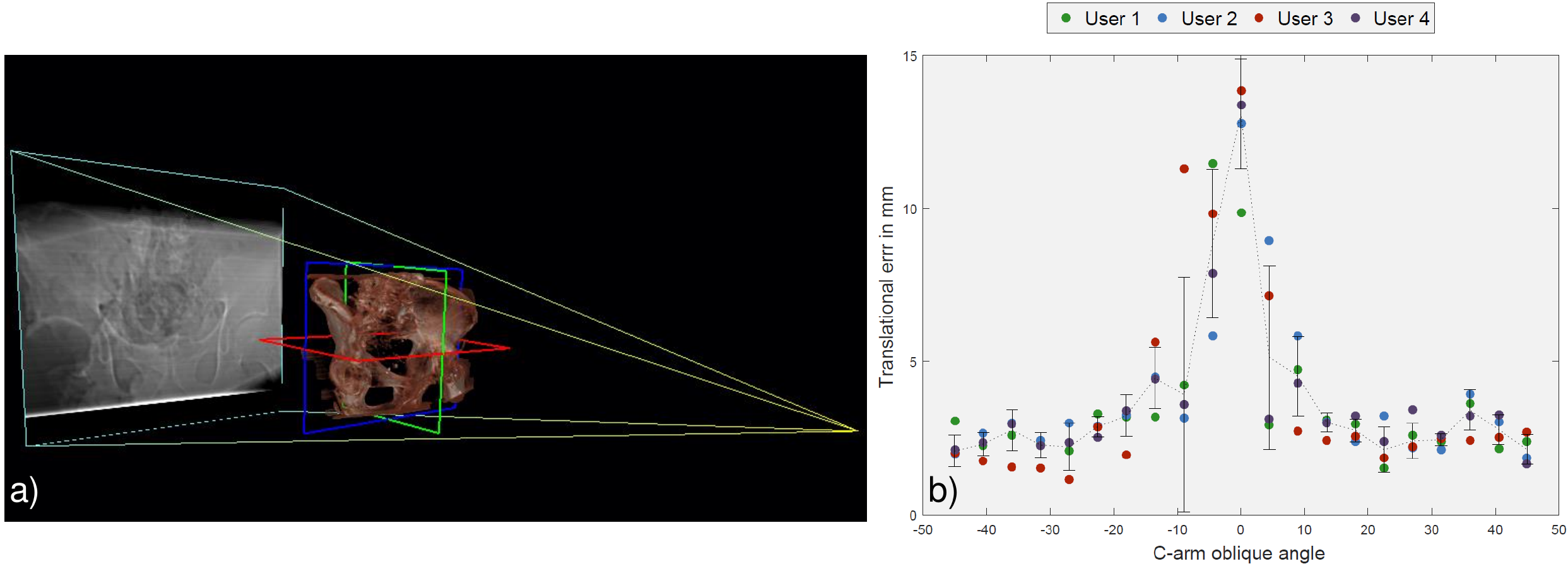}
  \caption{DRRs were generated from $-45^{\circ}$ to $+45^{\circ}$ around the AP view \textbf{(a)}. Participants were each time given two images, where one was always AP, and the other one generated from a different view. The translational errors are shown for all four participants in \textbf{(b)}. Note that $0^{\circ}$ in the horizontal axis refers to where the user performed planning on only the AP X-ray image.}
  \label{fig:DRRsError}
\end{figure*}

Four orthopedic surgery residents from the Johns Hopkins Hospital participate in the user study. The translation error in placing the cup are shown in Fig.~\ref{fig:DRRsError}. The abduction and anteversion errors are measured as zero as a result of presetting the desired angles. The abduction and anteversion adjusted by the user solely using AP image (classic DAA) are $6.52^{\circ} \pm 5.97^{\circ}$ and $1.82^{\circ} \pm 1.89^{\circ}$, respectively. Ground-truth for these statistics includes the 5 DOF pose of the cup in CT data (as the cup is a symmetric hemisphere, 1 DOF, i.\,e. rotation around the symmetry axis, is redundant), where abduction and anteversion angles are $40^{\circ}$ and $25^{\circ}$, respectively. 


\paragraph{Error evaluation in the AR environment:} To evaluate the agreement between surgeons' actions in the AR environment with their intra-operative planning, we measure the orientational error of the impactor after placement with respect to its planning. 

\begin{figure*}
  \centering
  \includegraphics[width=0.8\textwidth]{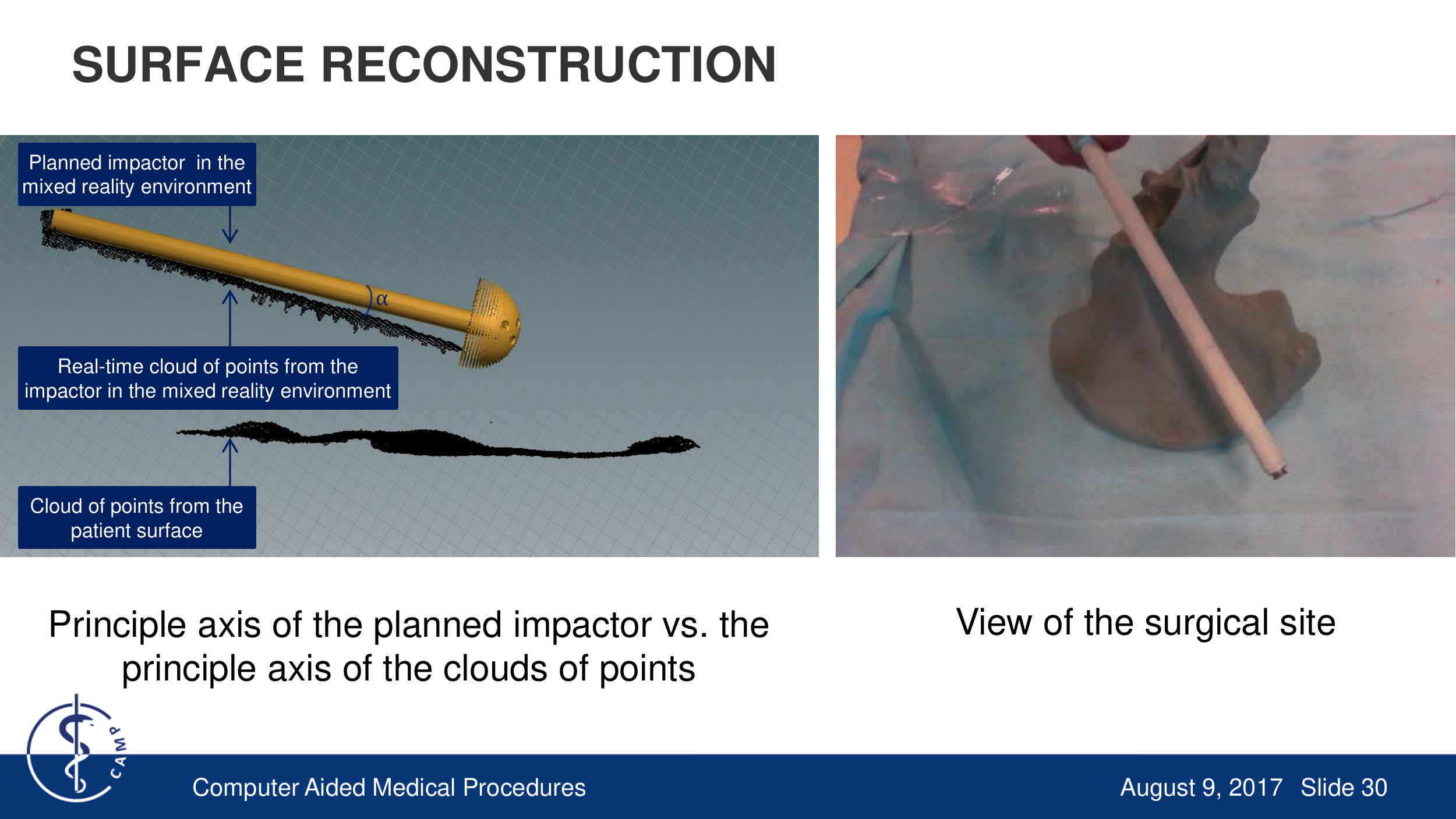}
  \caption{The angle between the principal axis of the virtual impactor and the cloud of points represent the orientation error in the AR environment.}
    \label{fig:principleaxis}
\end{figure*}

The axis-angle error between the principal axis of the true and planned impactor in the AR environment are measured as shown in Fig.~\ref{fig:principleaxis}. We repeat this experiment for 10 different poses, and each time use four virtual perspectives of the surgical site. The orientational error is $0.74^{\circ} \pm 0.41^{\circ}$.

After the cup is placed in the acetabulum using AR guidance, we acquire a CBCT scan of the cup and measure the translation, abduction, and anteversion errors compared to a ground-truth CBCT as $1.98$\,mm, $1.10^{\circ}$, and $0.53^{\circ}$, respectively.

\section{Discussion and Conclusion}\label{Sec::discuss}

We propose an AR solution based on intra-operative planning for easy and accurate placement of acetabular components during THA. Planning does not require pre-operative data, and is performed only on two stereo X-ray images. \textit{If either of the two X-ray projections is acquired from an AP perspective, the correct orientation of the cup can be adjusted automatically, thus, reducing the task load of the surgeon and promoting more accurate implant placement.}

Our AR environment is built upon an RGBD enhanced C-arm, that enables visualization of 3D optical information from the surgical site superimposed with the planning target. Ultimately, accurate cup placement is achieved by moving the impactor until it is fully aligned with the desired planning. 

Experimental results indicate that the anteversion and abduction errors are minimized substantially compared to the classic DAA approach. The translational error is below $3$\,mm provided that the lateral opening between two images is larger than $18^{\circ}$. All surgeons participating in the user study believed that presetting the cup orientation is useful and valid, as having access to AP images in the OR is a well-founded assumption. Nonetheless, the authors believe that a pose-aware RGBD augmented C-arm~\cite{fotouhi2017pose} can, in future, assist the surgeon in acquiring and confirming true AP images considering pelvis supine tilts in different planes. 

\todo{The translational and orientational error of the proposed AR solution is $1.98$\,mm and $1.22^{\circ}$ respectively which shows reduced error compared to navigation-based system proposed by Sato et.al. with translation error of $2.98$\,mm and orientation error of $4.25^{\circ}$~\cite{sato2000intraoperative}. These results show the clear necessity to continue research and perform user studies on cadaveric specimens and quantify the changes in operating time, number of required X-ray images, dose, accuracy, and surgical task load compared to classic image-guided approaches.}

In classic DAA hip arthroplasty, correct translation of the cup is achieved by naturally placing the acetabular component inside the acetabulum, and then moving the impactor around the pivot point of the acetabulum until the cup is at proper orientation. However, in order for our proposed solution to provide reliable guidance, both the translational and orientational alignments need to be planned. 

In addition to presetting the orientations for the cup during planning, the surgeon can also adjust all 6 DOF rigid parameters of the component. Though, in the suggested AR paradigm, there are two redundant degrees of freedom; $i)$ rotation along the symmetry axis of the cup, and $ii)$ translation along the acetabular axis. 

The RGBD camera on the C-arm is a short range camera to allow detection even in near distances. The RGB channel of the sensor is used for tracking visual markers, and the depth channel is utilized to enable AR. The field of view of the RGBD camera is greater than the X-ray camera. Therefore, it allows placing visual marker outside the X-ray view to not obscure the anatomy in the X-ray image. 

The visual marker is only introduced into the surgical scene for a short interval between acquiring two X-ray images. These external visual markers could be avoided if incorporating RGBD-based simultaneous localization and mapping to track the surgical site~\cite{fotouhi2017pose}. Alternatively, the impactor which is a cylindric object could be used as a fiducial for vision-based inside-out tracking. It is important to note that surgical tools with shiny surfaces reflect IR beam. Tracking the surgical impactor is only done reliably if the surface has a matte finish, or it is covered with a non-reflective adhesive material.

Projection of the 3D hemispheric virtual cup onto the plane of X-ray images are done by utilizing the intrinsics parameters of the X-ray camera. These parameters are estimated while performing the checkerboard calibration. However, at different C-arm arrangements the focal length and principal point could slightly change due to gravity and flex in the C-arm machine. \todo{We quantified the drift in the principal point for $\pm 10^{\circ}$, $\pm 20^{\circ}$, and $\pm 30 ^{\circ}$ of C-arm lateral opening and the average shift was $5.17$, $7.3$, and $17$ pixels on a $1024 \times 1024$ X-ray image. Considering the pixel spacing of the detector, these values are equivalent to $1.16$\,mm, $1.64$\,mm, and $3.82$\,mm drift on the detector plane coordinate frame.} To overcome the limitations of change of intrinsics in future, a look-up table could be constructed from pre-calibration of the C-arm at different angulations. The correct intrinsic parameters could then be retrieved from the table by matching the corresponding extrinsics from the inside-out tracking of the C-arm. \todo{To avoid small inaccuracies due to image distortion of the image intensifier, we placed the acetabulum near the image center where image distortion is minimal.}

\todo{During the clinical intervention, sterilization of the imaging device needs to be ensured by either covering the RGBD camera with transparent self-adhesive sterile covers, or extending the mount of the camera such that the camera is located outside the sterile zone. While both options are conceivable, the latter will reduce the range of free motion while rearranging the C-arm.}

\todo{The RGBD sensor is not embedded in the gantry of the C-arm; therefore, it is possible that the surgical crew inadvertently hit the camera and affect the calibration. Since repeating the co-calibration for the imaging devices is not feasible when the patient is present in the OR, we plan to place an additional co-calibrated camera on the opposite side of the detector. Hence, when the calibration of one camera becomes invalid, the opposite camera could be used as a substitute.}

\todo{In the proposed solution the patient is assumed to be static while placing the cup. However, if the patient moves, either the planning needs to be repeated, or the surgeon ought to continue with classic fluoroscopy-based guidance.}

This AR solution for THA uses a self-contained C-arm which only needs a one-time offline calibration, requires no external trackers, and does not depend on out-dated pre-operative patient data. We believe that this system, by enabling quick planning and visualization, can contribute to reduction of radiation, time, and frustration and increase the efficiency and accuracy for placing acetabular components. Ultimately, this approach may aid in reducing the risk of revision surgery in patients with diseased hip joints.

\end{spacing}
\end{document}